\documentclass[10pt,twocolumn,letterpaper]{article}

\usepackage{cvpr}
\usepackage{times}
\usepackage{epsfig}
\usepackage{graphicx}
\usepackage{amsmath}
\usepackage{amssymb}

\usepackage{multirow}
\usepackage[ruled,vlined]{algorithm2e}
\usepackage{algorithmic}
\usepackage{listings}
\usepackage{subfigure}
\usepackage[]{caption}
\usepackage{xcolor}
\usepackage[export]{adjustbox}
\usepackage{authblk}

\usepackage[pagebackref=true,breaklinks=true,letterpaper=true,colorlinks,bookmarks=false]{hyperref}

 \cvprfinalcopy 

\def\cvprPaperID{} 

\ifcvprfinal\fi
\begin{document}

\title{Few-Shot Meta-Denoising}

\author[1]{Leslie Casas}
\author[2]{Attila Klimmek}
\author[3]{Gustavo Carneiro}
\author[1]{Nassir Navab}
\author[2]{Vasileios Belagiannis}
\affil[1]{Technische Universit\"at M\"unchen, Garching bei M\"unchen, Germany}
\affil[2]{Universit\"at Ulm, Ulm, Germany}
\affil[3]{The University of Adelaide, Adelaide, Australia}
\affil[ ]{\normalsize \textit{firstname.lastname@\{tum.de, uni-ulm.de, adelaide.edu.au\}}}

\maketitle

\begin{abstract}
We study the problem of few-shot learning-based denoising where the training set contains just a handful of clean and noisy samples. A solution to mitigate the small training set issue is to pre-train a denoising model with small training sets containing pairs of clean and synthesized noisy signals, produced from empirical noise priors, and fine-tune on the available small training set. While such transfer learning seems effective, it may not generalize well because of the limited amount of training data. In this work, we propose a new meta-learning training approach for few-shot learning-based denoising problems. Our model is meta-trained using known synthetic noise models, and then fine-tuned with the small training set, with the real noise, as a few-shot learning task. Meta-learning from small training sets of synthetically generated data during meta-training enables us to not only generate an infinite number of training tasks, but also train a model to learn with small training sets -- both advantages have the potential to improve the generalisation of the denoising model. Our approach is empirically shown to produce more accurate denoising results than supervised learning and transfer learning in three denoising evaluations for images and 1-D signals. Interestingly, our study provides strong indications that meta-learning has the potential to become the main learning algorithm for denoising.
\end{abstract}

\section{Introduction}
\label{sec:introduction}

Learning-based denoising methods typically require a large training set with pairs of clean and noisy samples, where the simultaneous access to the pair is usually challenging. In this framework, training sets are often built using synthetic noise that is added to images based on a prior noise model~\cite{2010_vincent}. The use of prior noise models for training data generation models is also common in medical imaging, where the noise distribution is usually assumed to be known~\cite{yi2018sharpness, zeng2015simple}. In other cases, such as with motion or image signals~\cite{SIDD_2018_CVPR, 2016_burri}, it is possible to employ multiple sensors for simultaneous capturing of clean and noisy signals. However, this setup cannot be easily replicated in other real-world scenarios. Consequently, the availability of training sets that contain a large amount of pairs of clean and real noisy samples is limited, making the study of few-shot learning-based denoising an important research topic.

To frame our problem, we assume that: 1) we have access to clean samples (i.e.~noiseless), 2) we have empirical prior knowledge of the real noise distribution, and 3) we have a small training set containing pairs of clean and noisy samples affected by real noise. In this context, we introduce a new meta-learning training approach for few-shot learning-based denoising problems. In our proposed training approach, the meta-training stage consists of learning how to quickly denoise from a limited number of synthetically generated noisy samples. Afterwards, the fine-tuning stage comprises the learning from the limited training set containing pairs of clean and noisy samples affected by real noise. To the best of our knowledge, this paper is the first to propose a meta-learning method for few-shot learning in denoising problems.

Traditionally, signal denoising has been addressed by non-learning approaches~\cite{dabov2007image, 2011_madgwick, tagkey2009iii}, where solutions have only made the assumption (2) above. However,  learning-based works have shown more effective denoising results, when there is sufficient amount of training data~\cite{zhang2017beyond}. Denoising with supervised learning has been explored in three different ways. First, one can directly train a denoising model based on a small training set (assumption 3 above), but current deep learning models are likely to over-fit such small training sets. Second, the denoising model can be trained with supervision using samples formed from synthetic noise (assumptions 1 and 2), while it is tested on the real noise. The lack of fine-tuning on real noise typically results in low denoising accuracy~\cite{tzeng2015adapting}. Alternatively, in transfer learning, a denoising model is pre-trained with pairs of clean and synthetically generated noisy samples, where the noise is based on a known model (e.g.~Gaussian or Poisson). Then, this pre-trained model is fine-tuned with the small training set containing real noise. Unfortunately, depending on the size of the small training set with real noise, transfer learning may not generalize well~\cite{yosinski2014transferable}.

In this paper, we propose a new and effective meta-learning approach for learning-based few-shot denoising problems. Although we share the same assumptions of transfer learning, our paper is motivated by the success of meta-learning in few-shot classification~\cite{lake2013one}. The most successful few-shot approaches are based on meta-learning, where the model learns to quickly adapt to new learning tasks using small training sets~\cite{ravi2016optimization, snell2017prototypical}. We argue that the success of meta-learning is due to: 1) the ability to generate potentially infinite training sets containing synthetic noise (like transfer learning); and 2) the ability to train a model that quickly adapts to new learning tasks with small training sets (unlike transfer learning).
In our experiments, we empirically show that the denoising results produced by meta-learning are better than the ones from transfer learning, and other non-learning and learning-based denoising approaches.

We assess our work on three different data sets: low dose CT scans, natural images, and electrocardiogram (ECG) sequences. The demonstration of functionality in these diverse data sets shows the generalization ability of our proposed meta-denoising learning method, and allows us to compare it with relevant denoising methods. In all data sets, we rely on prior noise models for synthesizing noisy signals that partially capture the real noise affecting the image or signal. Our evaluations show that meta-learning outperforms supervised learning and transfer learning in few-shot denoising scenarios, given a sufficient number of learning tasks. Therefore, the main contribution of this paper is to be the first to show the potential of meta-learning to become a standard few-shot learning-based denoising algorithm.

\section{Related Work}

\paragraph{Traditional Denoising} This type of denoising has been performed with signal processing algorithms. On the image domain, it is common to assume that the real noise is related to the Gaussian and Poisson noise models~\cite{luisier2011image}. Similarly, the variance-stabilizing transformation has been proposed for converting the noise to Gaussian noise and then denoise with standard algorithms such as BM3D~\cite{dabov2007image} or~\cite{buades2005non}. In signal denoising, wavelets and filtering are the usual algorithms for denoising~\cite{2011_madgwick, 2006_sabatini, tagkey2009iii}. However, recent works have shown that learning-based approaches outperform such traditional denoising algorithms~\cite{SIDD_2018_CVPR, zhang2017beyond}.

\paragraph{Learning-based Denoising} Auto-encoders~\cite{vincent2008extracting, 2010_vincent} have standardized learning-based methods, followed by more complex network architectures~\cite{agostinelli2013adaptive, 2012_xie}. Convolutional neural networks~\cite{jain2009natural, rethage2018wavenet}, multi-layer perceptron~\cite{burger2012image} and encoder-decoder~\cite{brooks2019unprocessing, larsen2015autoencoding, 2016_mao, zhang2017beyond} architectures have also been proposed for image denoising. Unfortunately, since such methods are generally trained with supervision, they need large training sets of clean and noisy image pairs. However, as explained in Sec.~\ref{sec:introduction}, some denoising problems can have relatively small training sets and, consequently, be characterized as few-shot learning problem. Differently from previous learning-based denoising methods, our approach explicitly assumes that denoising is a few-shot learning problem.
An alternative to be considered for solving this few-shot learning problem is transfer learning~\cite{bengio2012deep, caruana1995learning}, where the denoising model is pre-trained in some related denoising tasks (e.g., using large-scale data sets that may be available or building large-scale data sets by synthesizing noisy signals or images).  Then, this pre-trained model is fine-tuned with the small training set. However, transfer learning does not usually generalize well from small data sets and requires substantial hyper-parameter tuning~\cite{yosinski2014transferable}. We show the transfer learning limitations in our evaluations, when compared with our meta-learning approach.

\paragraph{Meta-learning} The motivation for meta-learning is to train a model that is optimized to adapt to new learning tasks (sampled from a common, but latent, distribution of tasks), with relatively small training sets~\cite{hochreiter2001learning, naik1992meta, schmidhuber1992learning, thrun1996learning}. Meta-learning happens at two levels. First, the meta-learner obtains knowledge across multiple tasks (sampled from the latent distribution of tasks) in order to improve the base learner to adapt to new tasks. Second, the base learner acquires fast knowledge of a specific task using a small number of annotated data. Based on this scheme, several approaches have been proposed, which explore: metric learning~\cite{vinyals2016matching, koch2015siamese, snell2017prototypical}, memory augmentation (i.e.~model-based)~\cite{santoro2016meta, munkhdalai2017meta} and  optimization~\cite{finn2017model, ravi2016optimization,  nichol2018first}. 
Meta-learning approaches that explore optimization methods learn how to update the base learner with an iterative method~\cite{andrychowicz2016learning, hochreiter2001learning, li2016learning}. For example, the meta-learner in~\cite{ravi2016optimization} is an LSTM that learns the update rule for the base learner. Instead of the recurrent net, the meta-learner is modelled as batch stochastic gradient descent in model agnostic meta-learning (MAML)~\cite{finn2017model} and Reptile~\cite{nichol2018first}. Both approaches deliver state-of-the-art performance in few-shot classification. We have tried both algorithms for denoising and found that Reptile provides better training convergence and generalization than MAML. Furthermore, Reptile requires first-order gradients, which is computationally more efficient than MAML that works with second-order gradients~\cite{nichol2018first}. For those reasons, we focus on Reptile in our study of meta-learning few-shot denoising algorithms.

To the best of our knowledge, this paper is the first to adapt meta-learning for the few-shot learning-based denoising problem.

\section{Few-Shot Meta-Denoising Learning}

We reformulate the optimization-based meta-learning method Reptile~\cite{nichol2018first}, which has been originally developed for few-shot classification, to work for few-shot image and signal denoising. Our meta-denoising approach is trained in two stages: 1) during meta-training, the tasks consist of small sets of synthesized noisy samples that have been formed by taking clean samples and applying synthetic noise sampled from known models; and then 2) a small training set of real noisy samples is used to fine-tune the denoising model. Note that the two stages listed above allow us to access an infinite number of tasks for the meta-training stage that can help in the learning of a generic denoising model that quickly adapts to new denoising tasks that contain small training sets.

\subsection{Problem Definition}

We define the clean signal as $\mathbf{y} \in \mathbb R^D$ and the corrupted signal as $\mathbf{x} \in \mathbb R^D$, where this corruption occurs due to noise that is specific to the signal formation.
At each step of meta-learning, each clean signal is affected by a noise model, where task $\tau$ refers to the selected noise model and its parameters. For each task $\tau$, sampled from a latent distribution of tasks $p(\tau)$, we build a training set defined by $\{ \mathcal{X}_{\tau}, \mathcal{Y}_{\tau}\}$, where $\mathcal{X}_{\tau} = \{ \mathbf{x}_i \}_{i=1}^k$,  $\mathcal{Y}_{\tau} = \{ \mathbf{y}_i \}_{i=1}^k$, and $\mathbf{x}_i = h_{\tau}(\mathbf{y}_i)$, with $h_{\tau}: \mathbb R^D \rightarrow \mathbb R^D$ denoting the function that transforms the clean signal $\mathbf{y}_i$ into the noisy signal $\mathbf{x}_i$ by adding synthetic noise, according to the task $\tau$.  The data set containing real noise is represented by $\{ \mathcal{X}_{\tau^*},\mathcal{Y}_{\tau^*} \}$, where we assume that $\tau^* \sim p(\tau)$, and this set is divided into a training and a test set, denoted by $\{ \mathcal{X}^{(t)}_{\tau^*},\mathcal{Y}^{(t)}_{\tau^*} \}$ and $\{ \mathcal{X}^{(v)}_{\tau^*},\mathcal{Y}^{(v)}_{\tau^*} \}$, respectively.

\subsection{Denoising Objective}

The objective is to learn the parameters $\theta$ of a deep neural network that recovers the clean signal. The network is represented by $f_{\theta}:\mathbb R^D \rightarrow \mathbb R^D$ and is trained with back-propagation. Given a large training set of pairs of clean and noisy samples, the objective to minimize is
\begin{equation}
\mathcal{L}(\theta,\mathbf{x},\mathbf{y})=\mathbb{E}_{\mathbf{x},\mathbf{y}}[f_{\theta}( \mathbf{x})-\mathbf{y}]^{ 2 }.
\label{eq:LossFunc}
\end{equation}
This is the standard configuration for gradient-based supervised learning, which tends to work well when there is sufficient amount of training data. Since we assume that the available pairs of clean and noisy samples are limited, we define the problem of denoising within the context of few-shot meta-learning, which we refer to as \textit{meta-denoising}.

\subsection{Meta-denoising}

We define the \textit{base learner} to be the network $f_{\theta}(.)$. The learning process estimates the parameter $\theta$ that can be quickly adapted to denoising real signals, given a small training set, containing few pairs of clean and noisy signals. In the context of meta-denoising, a \textit{task} corresponds to denoising pairs of clean and synthesized signals, formed by applying a known noise model to the clean signals. 
Note that a task $\tau$ is formed not only by noise models, but also by noise parameters. Given a sampled task $\tau_{i} \sim p(\tau)$, Reptile performs $s$ steps of gradient descent update on the loss from~(\ref{eq:LossFunc}), described by:
\begin{equation}
\theta^{\prime}_{i} = g(\mathcal{L}(\theta,h_{\tau}(\mathbf{y}),\mathbf{y}), \theta, s) ,
\label{eq:InnerLoop}
\end{equation}
where $g(.)$ is the update operator that represents stochastic gradient descent or another optimization algorithm. This is the \textit{inner-loop} update. After iterating over $n$ tasks, the model parameters $\theta$ are adapted towards the new parameters, defined as:
\begin{equation}
\theta \leftarrow \theta + \epsilon \frac{1}{n}\sum_{i=1}^n(\theta^{\prime}_{i} - \theta),
\label{eq:TaskMin}
\end{equation}
where $\epsilon$ is the step-size. This is the \textit{outer loop} of the algorithm, where across tasks knowledge is acquired. Note that instead of being explicitly modeled, the \textit{meta-learner} is implicitly represented by the gradient-based learning in~(\ref{eq:TaskMin}). The number of iterations for the inner and outer loops and the batch-size of the inner loop compose the hyper-parameters of the learning algorithm. At the end of the learning process, the model is fine-tuned with $\{ \mathcal{X}^{(t)}_{\tau^*},\mathcal{Y}^{(t)}_{\tau^*} \}$ and tested with $\{ \mathcal{X}^{(v)}_{\tau^*},\mathcal{Y}^{(v)}_{\tau^*} \}$ -- this is the only step that involves annotated samples, i.e.~pairs of clean and noisy samples, where the noise is real. The complete learning and assessment processes are presented in Algorithm~\ref{algo}.

\begin{algorithm}[h]
\caption{Meta-Denoising Training.}\label{algo}
  \begin{algorithmic}[1]
  \STATE Initialize $\theta$
  \FOR{iteration = $1,2,\dots$}
  \FOR{task = $1,2,\dots$, n}
      \STATE Sample noise model $\tau \sim p(\tau)$
      \STATE Sample k clean samples $\mathcal{Y}_{\tau}=\{ \mathbf{y}_i\}_{i=1}^k$ 
      \STATE Generate k noisy samples  $\mathcal{X}_{\tau}=\{ h_{\tau}(\mathbf{y}_i) \}_{i=1}^k$
     \STATE Compute $\theta^{\prime}_{task} = g(\mathcal{L}(\theta,\mathcal{X}_{\tau},\mathcal{Y}_{\tau}, \theta, s)$
  \ENDFOR
  \STATE Update $\theta \leftarrow \theta + \epsilon \frac{1}{n}\sum_{task=1}^n(\theta^{\prime}_{task} - \theta)$
 \ENDFOR
 \STATE Train $f_{\theta}(\mathbf{x})$ with $\{ \mathcal{X}^{(t)}_{\tau^*},\mathcal{Y}^{(t)}_{\tau^*} \}$
 \STATE Test $f_{\theta}(\mathbf{x})$ with $\{ \mathcal{X}^{(v)}_{\tau^*},\mathcal{Y}^{(v)}_{\tau^*} \}$.
 \end{algorithmic}
 \end{algorithm}

\begin{table}
\begin{tabular}{|l|c|c|c|c|}
\hline
                              & \multicolumn{2}{c|}{\# Tasks}                            & \multicolumn{2}{c|}{Tube current} \\ \hline
                              & Poisson                     & Gaussian                     & 10\%   & 5\%   \\ \hline
Initial Noise                &            -               &             -                &  38.28      &    35.18    \\ \hline
\multirow{5}{*}{\shortstack[l]{Supervised\\ Learning}} &                 4         &             0        & 36.50  & 34.15  \\ \cline{2-5} 
                              &                 4         &             4           & 35.50  & 33.80      \\ \cline{2-5}
                              &                 8         &             0           & 36.31  & 33.94    \\ \cline{2-5} 
                              &                 8         &             8           & 35.01  & 33.42    \\ \hline
\multirow{5}{*}{\shortstack[l]{Transfer\\ Learning}} &                 4         &             0           & 38.98  & 36.96  \\ \cline{2-5} 
                              &                 4         &             4         & 37.91  & 35.99 \\ \cline{2-5}
                              &                 8         &             0         & 38.39  & 36.70\\ \cline{2-5} 
                              &                 8         &             8         & 39.19  & 36.76\\ \hline
\multirow{5}{*}{\shortstack[l]{Meta-Denoising}} &                 4         &             0         & 39.77  & 37.37\\ \cline{2-5} 
                              &                 4         &             4         & 39.42  & 37.50\\ \cline{2-5}
                              &                 8         &             0         & \textbf{39.99}  & \textbf{37.96}\\ \cline{2-5}
                              &                 8         &             8         & 39.84  & 37.49\\  \hline
\end{tabular}
\caption{CT-Scan Evaluation. We examine two types of noise models, namely Gaussian and Poisson. We present the results for Gaussian and / or Poisson noise models. We evaluate on tube current with dose $10\%$ and $5\%$. The evaluation metric is the peak signal-to-noise ratio (PSNR), where larger is better. Bold numbers indicate the best performance per column.}
\label{eval-CT2}
\end{table}

\begin{table}[]
\begin{tabular}{|l|c|c|c|c|}
\hline
                            & \multicolumn{2}{c|}{Tube current} & Execution (sec.)\\\hline
                            & 10\%   & 5\%   &\\ \hline
Initial Noise               &  38.28      &    35.18   & -  \\ \hline
BM3D                        &  38.70&    36,27  & 2.42 \\ \hline
LGP-PCA                    &  38.98      &    36,29  & 206.84 \\ \hline
\shortstack[l]{Supervised\\Learning}        &  36.50      &    34.15 &  \textbf{0.0008} \\ \hline
\shortstack[l]{Transfer\\Learning}                   &  39.19      &    36.96 & \textbf{0.0008}  \\ \hline
\shortstack[l]{Meta-Denoising}                    &  \textbf{39.99}      &    \textbf{37.96} &  \textbf{0.0008} \\ \hline
\end{tabular}
\caption{Comparison with Non-learning Algorithms on CT-Scan. We compare the three learning-based algorithms with two classic denoising approaches (BM3D and LGP-PCA). The evaluation metric is the peak signal-to-noise ratio (PSNR), where larger is better.  The execution time is shown in seconds. Bold numbers indicate the best performance per column.}
\label{eval-CT3}
\end{table}

\section{Experiments}

We compare the proposed meta-denoising few-shot learning algorithm with other non-learning and learning-based methods. We assess our few-shot meta-denoising approach in two image-based datasets and one 1-D dataset to demonstrate the generalization of our approach to different types of signals. Since this paper is the first to explore few-shot denoising with meta-learning, we propose a new assessment protocol.

\noindent \textbf{Protocol.} For meta-training, noise models are sampled and synthetic tasks are generated as a k-shot denoising problem, where k-pairs of clean and noisy samples are produced. During meta-testing, real tasks are used for k-shot fine-tuning. The evaluation is performed on the test set, which is composed of previously unseen pairs of clean and noisy samples from the real tasks. The evaluation metric is the peak signal-to-noise ratio (PSNR) for images and signal-to-noise ratio (SNR) for 1-D data. These are the standard metrics for evaluating denoising methods~\cite{kabir2012denoising,2012_xie}.

\noindent \textbf{Other Training Algorithms.} We define two baselines for comparison with our approach. First, we train from scratch a \textit{supervised learning} model, where the training set is composed of synthetically generated data and is as large as the training set of all tasks in meta-training. As a result, the amount of data processed by supervised learning is the same as by meta-training.
Second, we perform \textit{transfer learning} with the trained model using the same k-pairs of clean and real noisy samples that we employ for fine-tuning in meta-training. By fixing the model and data, we can compare meta-learning with supervised- and transfer-learning in a fair manner. Finally, we also report the \textit{initial noise} for each evaluation that corresponds to the test set noise prior to applying a denoising algorithm.

\noindent \textbf{Implementation.} We present three experiments with different data types, so we employ different network architectures, optimization methods and hyper-parameter values. All employed architectures are taken from related publications, since proposing a new architecture is not the aim of this paper. The hyper-parameters have been chosen with grid-search optimization. The implementation of our approach, as well as the evaluation protocol will be made publicly available -- both implementations were made using PyTorch~\cite{paszke2017automatic}.

Below, for each data set, we discuss the task generation, implementation details and denoising results. Finally, we examine the results significance using the t-test, different k-shots and we also provide results when training with the MAML algorithm.

\subsection{CT-Scan Denoising Evaluation}\label{sec:CTEval}

\setlength{\fboxsep}{0pt}%
\setlength{\fboxrule}{1pt}%
\begin{figure*}[]
\centering
\begin{minipage}{.9\textwidth}
\centering
{\label{figCT1:a}\includegraphics[width=45.0mm]{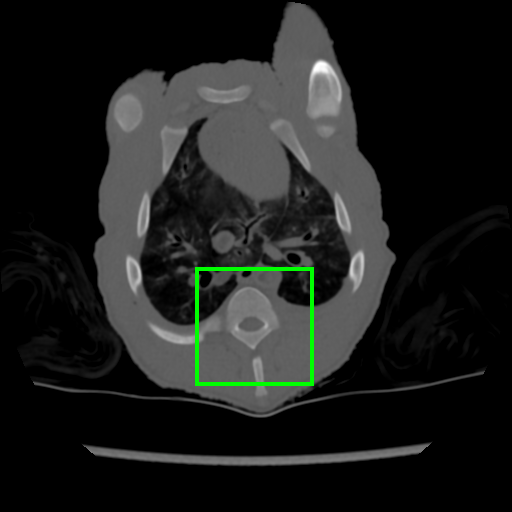}\llap{\includegraphics[height=1.5cm,cfbox=green 1pt]{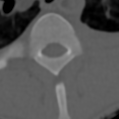}}}
{\label{figCT1:b}\includegraphics[width=45.0mm]{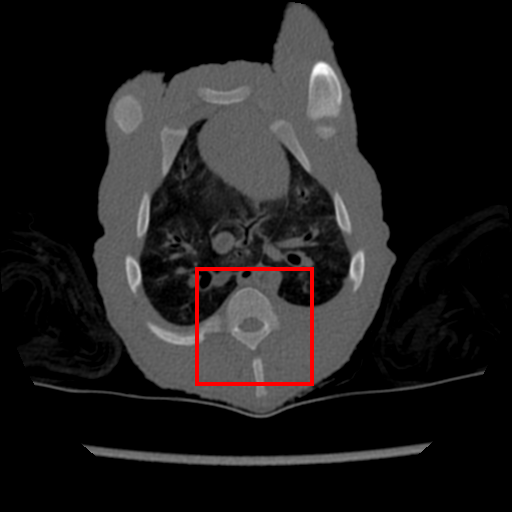}\llap{\includegraphics[height=1.5cm,cfbox=red 1pt]{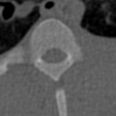}}}
{\label{figCT1:c}\includegraphics[width=45.0mm]{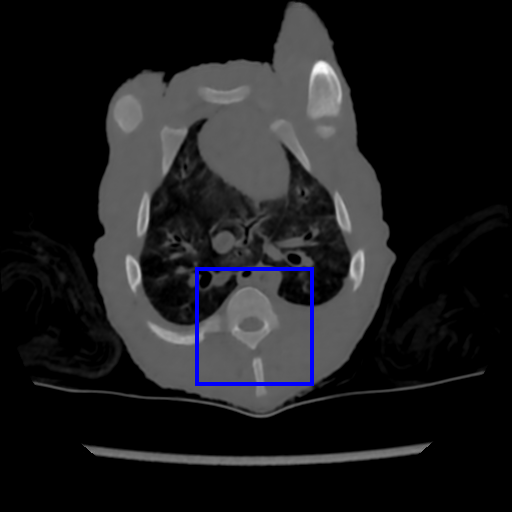}\llap{\includegraphics[height=1.5cm,cfbox=blue 1pt]{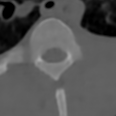}}}\\
\subfigure[Ground-truth.]{\label{figCT2:a}\includegraphics[width=45.0mm]{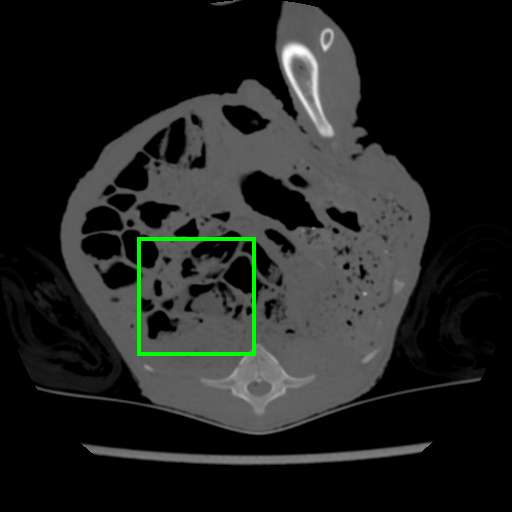}\llap{\includegraphics[height=1.5cm,cfbox=green 1pt]{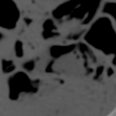}}}
\subfigure[Noisy Input.]{\label{figCT2:b}\includegraphics[width=45.0mm]{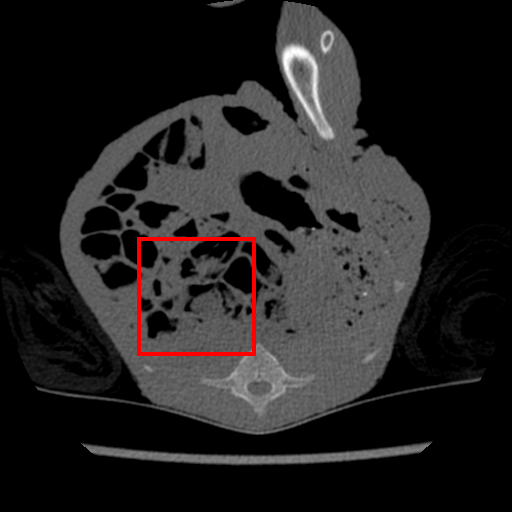}\llap{\includegraphics[height=1.5cm,cfbox=red 1pt]{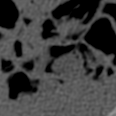}}}
\subfigure[Denoised Output.]{\label{figCT2:c}\includegraphics[width=45.0mm]{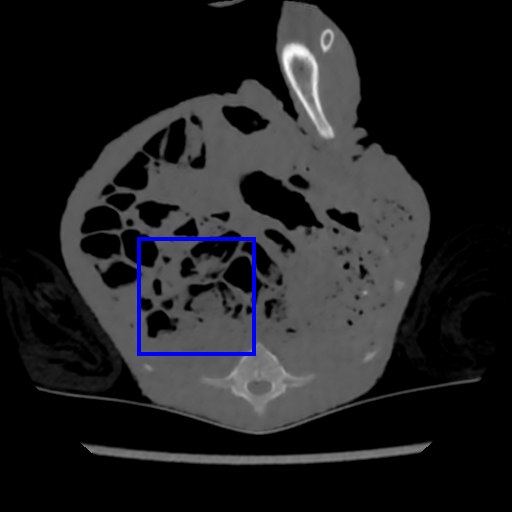}\llap{\includegraphics[height=1.5cm,cfbox=blue 1pt]{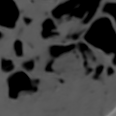}}}
\caption{CT-Scan Denoising. We show two examples of results (top and bottom rows) on CT-Scan~\cite{yi2018sharpness} denoising dataset, where the columns (from left to right) show the clean image, noisy image, and denoised image produced by our approach. We also show a magnified visualization for each case.}
\label{fig:visualResCT}
\end{minipage}
\end{figure*}

\noindent \textbf{Task Generation.} The CT database in~\cite{yi2018sharpness} contains CT scans of a deceased piglet. The scans are produced using a normal dose of tube current (100\%) and reduced doses. We use the doses of 10\% and 5\% because they include substantial amount of noise, where each dose set has 850 images of size $D=512 \times 512$. Following the data set protocol~\cite{yi2018sharpness}, 708 of those images are selected for training and 108 for testing. During training, each task consists of synthetic images generated by simulating reduced dose CT images by adding Poisson and Gaussian noise to the sinograms of the normal dose CT images, as in $\mathbf{x} = h_{\tau = \mathcal{P}(b),\mathcal{N}(0,\sigma^2)}(\mathbf{y})$, where $\mathbf{x}  = S^{-1}(\mathbf{z})$, $\mathbf{z} \sim \mathcal{P}(b \times \exp(-S(\mathbf{y})))+\mathbf{r}$, with $\mathcal{P}(.)$ denoting the Poisson distribution, $S(.)$ representing the process to generate the sinogram or the fan-beam projected data~\cite{zeng2015simple} from the clean CT image $\mathbf{y}$, $\mathbf{r} \sim \mathcal{N}(0,1)$ denoting the read-out electronic noise represented by $\mathcal{N}(.)$ the Gaussian distribution with mean $\mu$ and variance $\sigma$, and $b$ denoting the X-ray source intensity or a blank scan factor that controls the noise level. We create the tasks with $\mathbf{r}=0$, namely one per blank scan factor, where we use the following factors $b \in \{{10}^{4},{3\times10}^{4},{5\times10}^{4},{7\times10}^{4},{10}^{5},{2\times10}^{5}, {3\times10}^{5},$ ${5\times10}^{5}\}$. Consequently, each factor contributes to the generation of a different task. To simulate the effect of the Gaussian noise, we add noise directly on the CT image $\mathbf{x}$. The Gaussian task $\tau$ is represented by Gaussian noise $\eta(i,j) \sim \mathcal{N}(0, \sigma^{2})$ with $\sigma^2 \sim U(0.001,0.003)$, where $U(a,b)$ denoting uniform distribution in $[a,b]$. The result is the noisy image $\mathbf{x}(i,j) = \mathbf{y}(i,j) + \eta(i,j)$ for $i\in \{1,...,D_{height}\}$ and $j\in \{1,...,D_{width}\}$. Finally, we keep $k$ images for the meta-learning for each reduced dose set. The rest of the images are used for evaluation.

\noindent \textbf{Model.} We use a residual encoder-decoder architecture from~\cite{chen2017low} that has been proposed for low dose CT denoising~\cite{yi2018sharpness}. The network consists of a 10-layer convolutional neural network with input and output sizes of $55 \times 55$. The activation functions are all ReLU, except for the last layer that is linear. The optimization algorithm is Adam~\cite{kingma2014adam} with learning rate $10^{-4}$. The inner loop has 20 epochs with mini-batch size 20. The outer loop runs for 1000 epochs with step-size $\epsilon = 0.1$ and the number of fine-tuning samples $k=10$.

\noindent \textbf{Results.} We evaluate CT-Scan denoising on tube current with doses $10\%$ and $5\%$, where smaller percentage denotes more noise -- results are summarized in Table~\ref{eval-CT2}. We generate from 4 to 8 Gaussian and Poisson tasks, with each task representing an individual noise model. Notice that we also applied Gaussian and Poisson noise, at the same time, on an image as another task, but the performance was not as good as sampling independently. In the case of task combination (e.g.~4 Gaussian and 4 Poisson), we equally sample tasks from both noise models. We attempted to generate less than 4 and more than 8 tasks, but there was not a considerable performance difference.

The same evaluation is performed for the supervised- and transfer-learning. Comparing with these two algorithms, it is clear that our approach delivers the highest PSNR. Although the results from transfer learning are closer, the difference is significant based on the t-test as we show in Sec.~\ref{sec:statistical_significance}. Since this is the only experiment where we have access to a relatively large training set containing pairs of clean and real noisy  data, we check the upper bound performance for our approach, which is the supervised learning performance using this large training set. Note that such full supervision violates one of our three assumptions mentioned in Sec.~\ref{sec:introduction} -- in particular, the one regarding the availability of a large number of training samples. Supervised learning achieves 40.28dB for $10\%$ and 38.84dB for $5\%$, when trained with clean and real noisy training data. Our results in Table~\ref{eval-CT2} indicate that we are close to this upper bound, although we train with synthetic training data. To further explore our approach, we meta-train it with clean and real noisy training data. Notably, our meta-denoising reaches 40.47dB for $10\%$ and 38.31dB for $5\%$. This result shows the potential of meta-learning to become the standard training algorithm for the task of denoising. We present visual results in Fig.~\ref{fig:visualResCT}.

Furthermore, we investigate how our method and the other learning-based algorithms from Table~\ref{eval-CT2} compare to non-learning denoising algorithms. In Table~\ref{eval-CT3}, we present the best results from all learning-based approaches and compare with the non-learning approaches BM3D~\cite{dabov2007image} and LGP-PCA~\cite{zhang2010two}. It is clear that our meta-denoising, as well as, supervised- and transfer-learning outperform the non-learning approaches. Moreover, BM3D and in particular LGP-PCA are significantly slower in terms of execution time.

\subsection{Natural Images Evaluation}

\noindent \textbf{Task Generation.} We use the color data set BSD500~\cite{arbelaez2011contour} as training set and evaluate on the SIDD database~\cite{SIDD_2018_CVPR}. BSD500 data set is standard for image denoising~\cite{2016_mao,zhang2017beyond}, while the SIDD has been recently introduced for denoising images from smartphones. The training and testing sets from BSD500 have 200 images which we employ as training data during meta-training. SIDD has 160 images from which we keep $10$ for the fine-tuning stage and the rest as test set. In BSD500, the size of images is $D = 481 \times 321 \times 3$. In the training images, we add Gaussian noise in the spatial domain to generate tasks with synthetic noise, as in $\mathbf{x} = h_{\tau = \mathcal{N}(\mu,\sigma)}(\mathbf{y})$. This Gaussian noise has mean $\mu = 0$ and standard deviation $\sigma \in \{15,25,50\}$, which have been chosen from the literature in image denoising~\cite{2012_xie, zhang2017beyond}). Second, we introduce Poisson noise in the image domain as in $\mathbf{x} = h_{\tau = \mathcal{P}(b)}(\mathbf{y})$, where $\mathbf{x} \sim \mathcal{P}(b\times 10^{12} \times \mathbf{x})/10^{12}$, with $\mathcal{P}(.)$ denoting the Poisson distribution, and $b \in \{10^{9},10^{9.5},10^{10}\}$ denotes the scale factor that controls the noise level. In total, we have three tasks for Poisson and three for Gaussian noise.

\noindent \textbf{Model.} The network is DnCNN~\cite{zhang2017beyond}, an encoder-decoder ConvNet that reconstructs the residual noisy image. We extend the original architecture from~\cite{zhang2017beyond} to have a depth of $13$ layers to reduce the amount of resources required. The denoised signal is obtained by subtracting the predicted noise from the input noisy image. All layers of this network consists of convolutions, with batch normalization and ReLU activation, except the last layer, which is linearly activated. The input and output resolution image resolution is $55 \times 55$ during training, while $40 \times 40$ in testing. In addition, we evaluate a more recent version of the architecture DnCNN, xDnCNN~\cite{kligvasser2018xunit} which uses the new activation function xUnit adapted for image restoration scenarios. The optimizer for the inner loop is Adam with learning rate $10^{-3}$, with 20 inner epochs. The step-size of the outer loop is $\epsilon = 1e-4$ and the number of samples $k = 10$.

\noindent \textbf{Results.} Although the noise of natural images is more complicated than the proposed Gaussian and Poisson tasks, we are able to denoise images with real noise~\cite{SIDD_2018_CVPR} -- Table~\ref{eval-Natural} summarizes the results. We have used three Gaussian and Poisson tasks jointly, but independently. The results in Table~\ref{eval-Natural} indicate that the supervised learning performs poorly because the generated tasks may not represent well the real noise in natural images. However, the tasks are enough for meta-learning, as shown by the results.

\begin{figure*}[]
\centering
\begin{minipage}{.9\textwidth}
\centering
{\label{figNI3:a}\includegraphics[width=46.0mm]{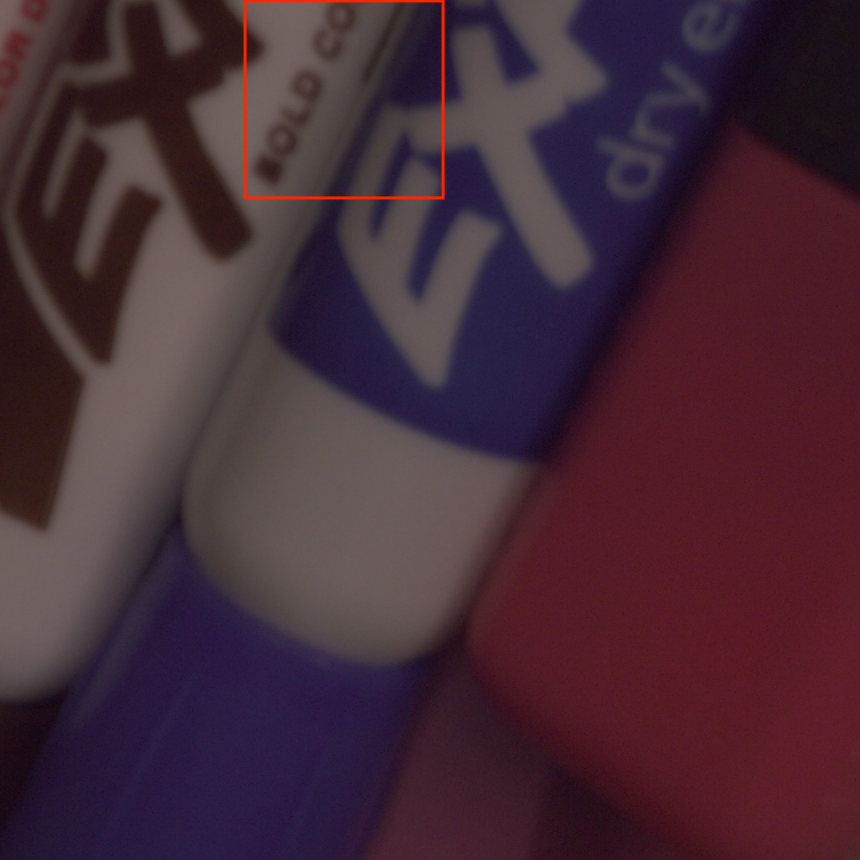}\llap{\includegraphics[height=1.7cm,cfbox=red 1pt]{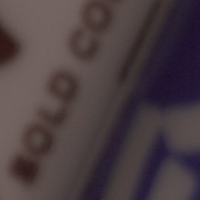}}}
{\label{figNI3:b}\includegraphics[width=46.0mm]{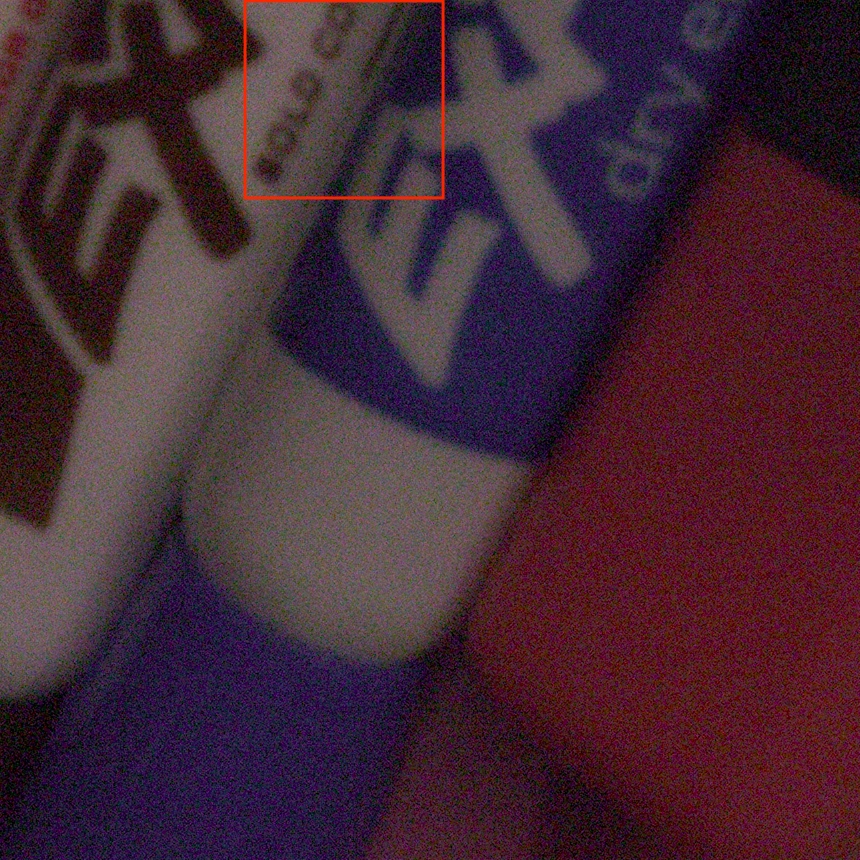}\llap{\includegraphics[height=1.7cm,cfbox=red 1pt]{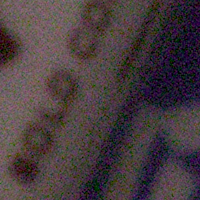}}}
{\label{figNI3:c}\includegraphics[width=46.0mm]{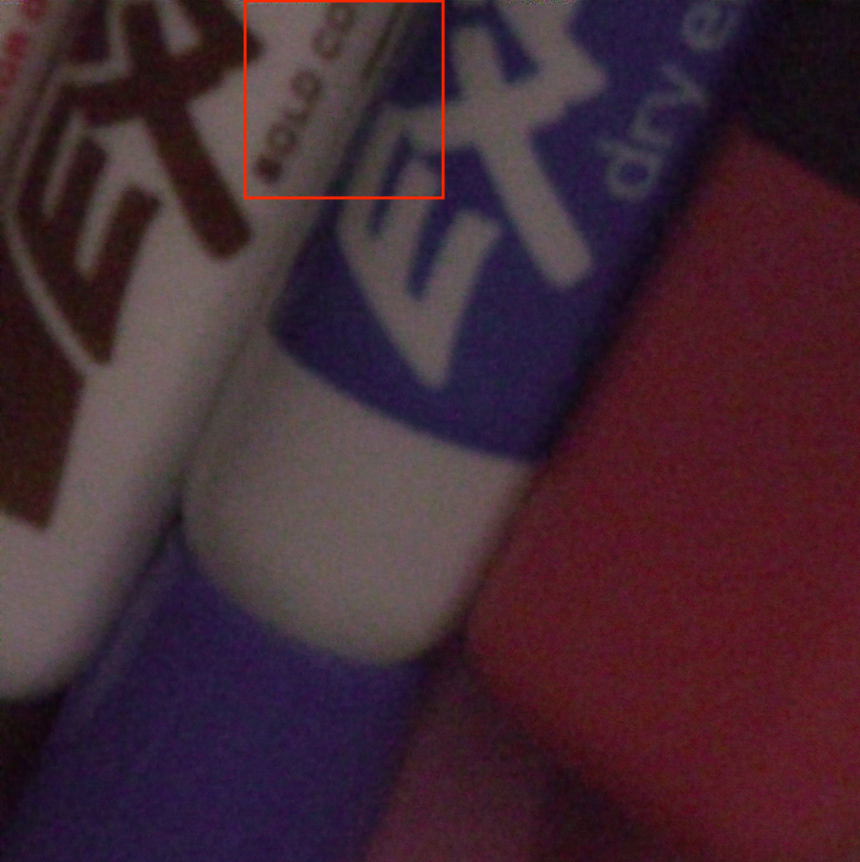}\llap{\includegraphics[height=1.7cm,cfbox=red 1pt]{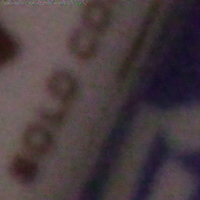}}}
\subfigure[Ground-truth.]{\label{figNI2:a}\includegraphics[width=46.0mm]{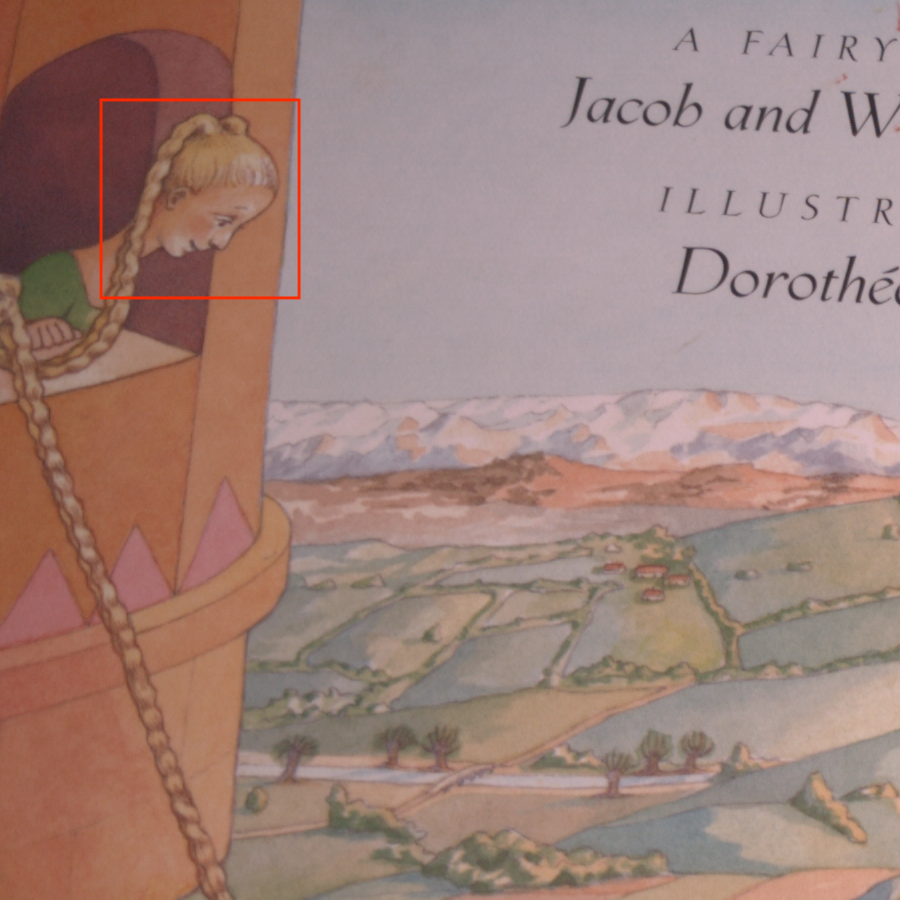}\llap{\includegraphics[height=1.7cm,cfbox=red 1pt]{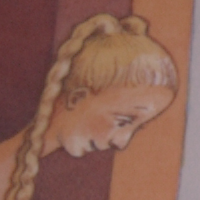}}}
\subfigure[Noisy Input.]{\label{figNI2:b}\includegraphics[width=46.0mm]{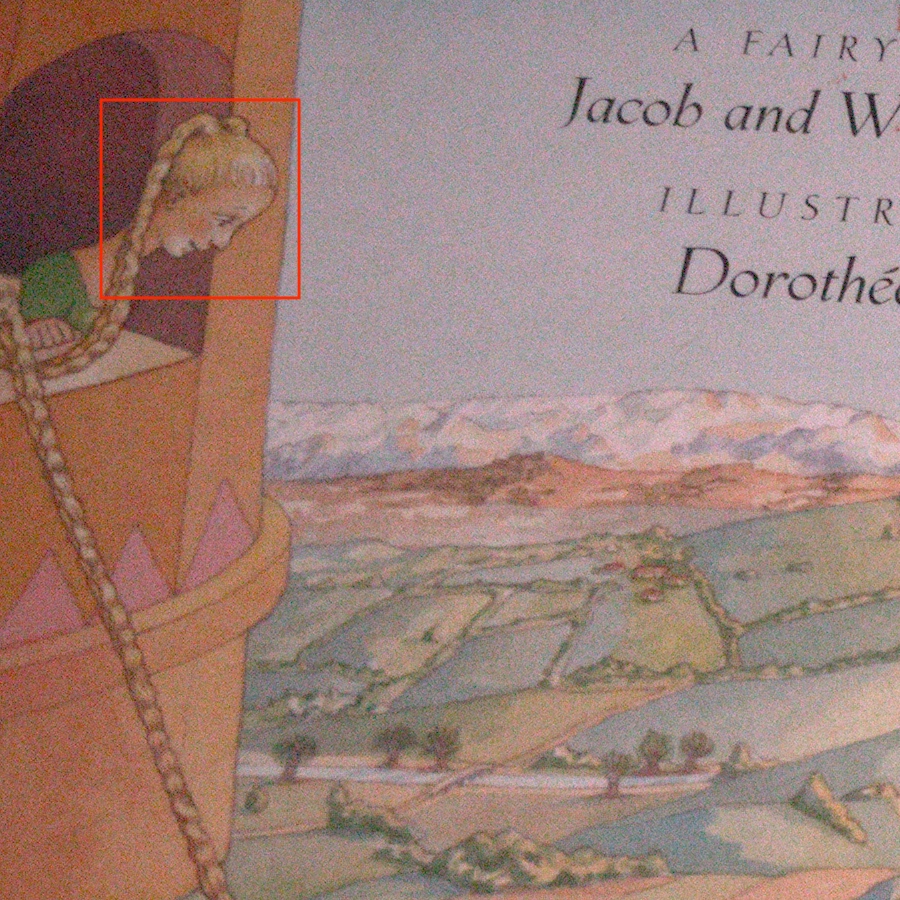}\llap{\includegraphics[height=1.7cm,cfbox=red 1pt]{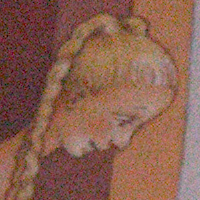}}}
\subfigure[Denoised Output.]{\label{figNI2:c}\includegraphics[width=46.0mm]{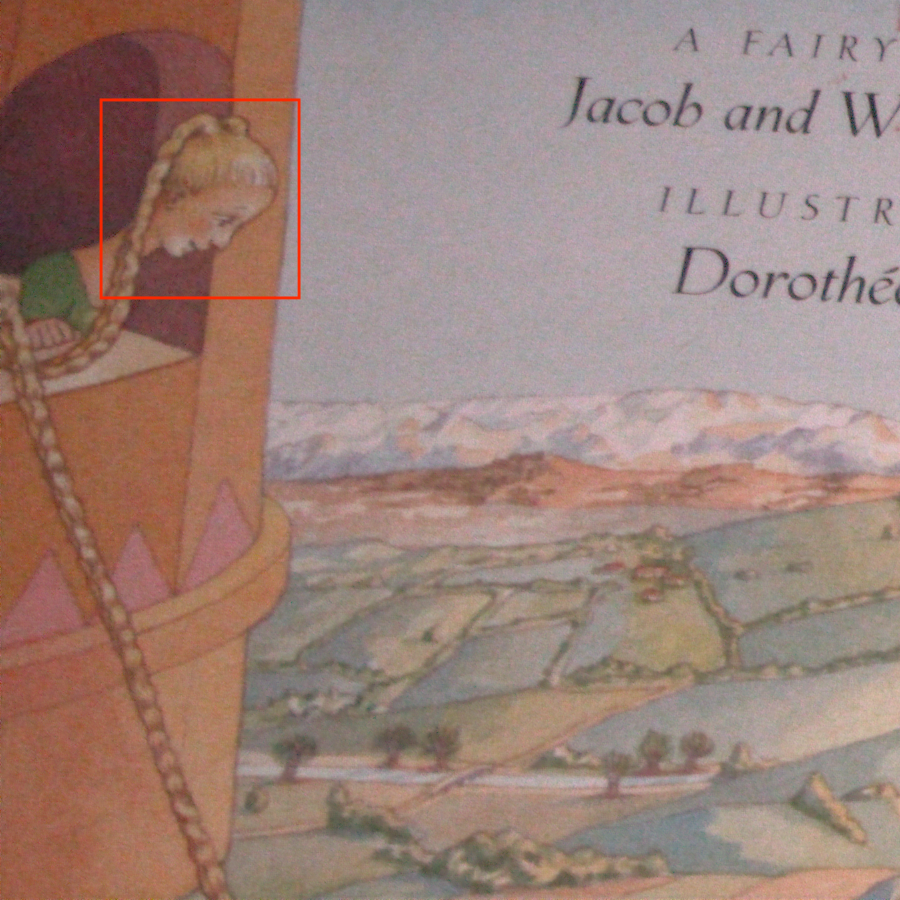}\llap{\includegraphics[height=1.7cm,cfbox=red 1pt]{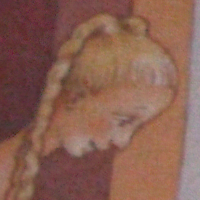}}}
\caption{Natural Image Results. We show two examples of results (top and bottom rows) on natural image~\cite{SIDD_2018_CVPR} denoising, where the columns (from left to right) show the clean image, noisy image, and denoised image produced by our approach. We also provide magnified visualization for each case.}
\label{fig:visualResNI}
\end{minipage}
\end{figure*}

\begin{figure*}[h]
	\centering
	\includegraphics[height=3.2cm]{imgc_59_gt.png}
	\includegraphics[height=3.2cm]{imgc_59_in.png}
	\includegraphics[height=3.2cm]{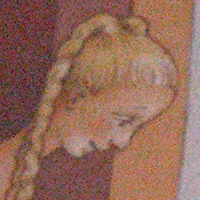}
	\includegraphics[height=3.2cm]{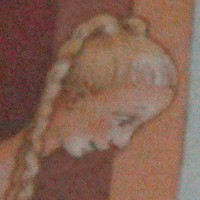}
	\includegraphics[height=3.2cm]{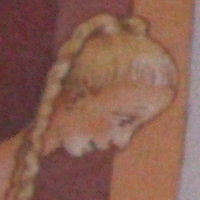}\\
	
	\includegraphics[height=3.2cm]{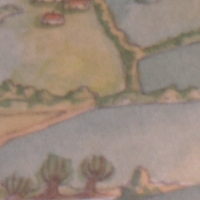}
	\includegraphics[height=3.2cm]{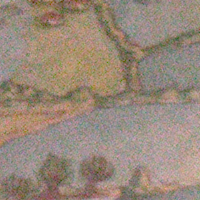}
	\includegraphics[height=3.2cm]{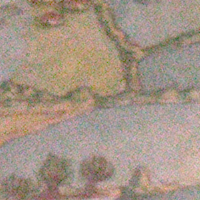}
	\includegraphics[height=3.2cm]{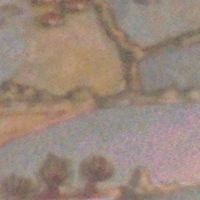}
	\includegraphics[height=3.2cm]{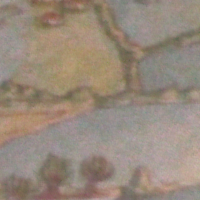}\\

	\includegraphics[height=3.2cm]{imgc_125_gt.png}
	\includegraphics[height=3.2cm]{imgc_125_in.png}
	\includegraphics[height=3.2cm]{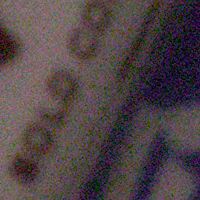}
	\includegraphics[height=3.2cm]{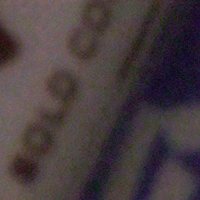}
	\includegraphics[height=3.2cm]{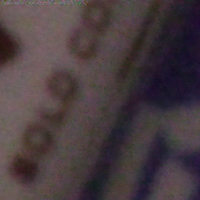}\\
	
	\includegraphics[height=3.2cm]{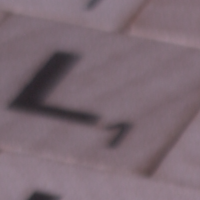}
	\includegraphics[height=3.2cm]{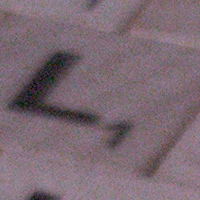}
	\includegraphics[height=3.2cm]{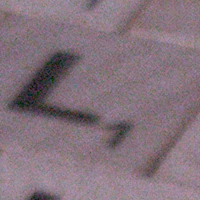}
	\includegraphics[height=3.2cm]{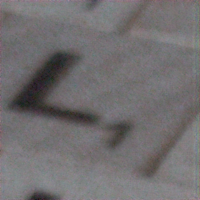}
	\includegraphics[height=3.2cm]{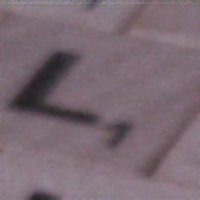}\\
	
	\subfigure[c][\textbf{Ground-truth.}]{\includegraphics[height=3.2cm]{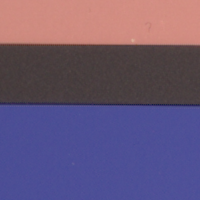}}
	\subfigure[c][\textbf{Noisy Input.}]{\includegraphics[height=3.2cm]{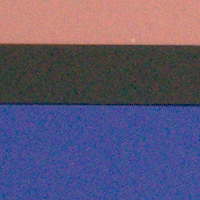}}
	\subfigure[c][\textbf{Supervised Learning.}]{\includegraphics[height=3.2cm]{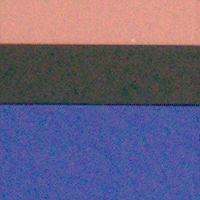}}
	\subfigure[c][\textbf{Transfer Learning.}]{\includegraphics[height=3.2cm]{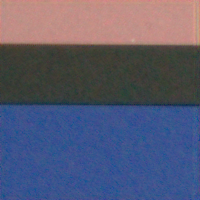}}
	\subfigure[c][\textbf{Meta-Learning (Ours).}]{\includegraphics[height=3.2cm]{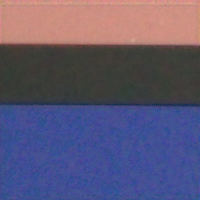}}
	\caption{More Natural Image Results. We provide additional image patches with visual comparison of supervised learning (c), transfer learning (d) and our approach based on meta-learning (e).}
	\label{fig:visualResComp}
\end{figure*}

Similarly to the CT-Scans, we compare all learning-based algorithms to non-learning denoising approaches. In Table~\ref{eval-Natural1}, we present the best results of meta-denoising and the other two learning algorithms (from Table \ref{eval-Natural}), next to BM3D~\cite{dabov2007image} and LGP-PCA~\cite{zhang2010two}. In this evaluation, the non-learning algorithms perform considerably better at the cost of a significantly longer execution time. This finding suggests a limitation of learning-based denoising methods, which is that they depend on training sets that represent well the noise seen in the evaluation data set. A few visual results of our approach are presented in Fig.~\ref{fig:visualResNI} and Fig.~\ref{fig:visualResComp}.

\begin{table}[]
\centering
\resizebox{\columnwidth}{!}{%
\begin{tabular}{|l|c|c|c|c|c|}
\hline
                                & \multicolumn{2}{c|}{\# Tasks}   &   \multicolumn{2}{c|}{PSNR}  \\\hline
                                & Gaussian  & Poisson         &  DnCNN  & xDnCNN              \\\hline        
Initial Noise                   & -         &     -                     & \multicolumn{2}{c|}{27.61} \\ \hline
\multirow{2}{*}{\shortstack[l]{Supervised\\ Learning}} & 3 &   0        & 22.23 & 26.71         \\ \cline{2-5} 
                                &  3        & 3                         & 25.74 & 26.25          \\ \hline
\multirow{2}{*}{\shortstack[l]{Transfer\\ Learning}} & 3 &   0         & 28.54 & 29.11            \\ \cline{2-5} 
                              &  3          &3                          & 30.01 & 28.30          \\ \hline
\multirow{2}{*}{\shortstack[l]{Meta-Denoising}} & 3 &    0      & 29.88 & \textbf{29.58}         \\ \cline{2-5} 
                              & 3           &3                          & \textbf{31.71} & 29.51          \\ \hline                  
\end{tabular}}
\caption{Natural Image Evaluation. We evaluate on the smartphone dataset~\cite{SIDD_2018_CVPR} our meta-denoising approach, supervised- and transfer- learning. The evaluation metric is the peak signal-to-noise ratio (PSNR) and tasks are generated from a Gaussian and Poisson distributions. The best result per column is in bold font.}
\label{eval-Natural}
\end{table}

\begin{table}[]
\centering
\begin{tabular}{|l|c|c|c|c|c|}
\hline
                                &    PSNR& Execution (sec.)  \\\hline
Initial Noise                   & 27.61 & - \\ \hline
BM3D                           &  \textbf{34.21}     &  273.7 \\ \hline
LGP-PCA                         &  32.41     & 11009  \\ \hline
\shortstack[l]{Supervised\\ Learning} & 26.71 &    \textbf{0.001}      \\ \hline
\shortstack[l]{Transfer\\ Learning}  & 30.01 &       \textbf{0.001}     \\ \hline 
\shortstack[l]{Meta-Denoising} & 31.71 &  \textbf{0.001}        \\ \hline 
\end{tabular}
\caption{Comparison with Non-learning Algorithms on Natural Images. We compare the three learning-based algorithms with two classic non-learning denoising approaches (BM3D and LGP-PCA). The execution time in seconds is reported too. The best result per column is in bold font.}
\label{eval-Natural1}
\end{table}

\subsection{ECG Denoising Evaluation}

\noindent \textbf{Task Generation.} We use the Physionet ECG-ID database~\cite{2000_physionet}, which includes 310 ECG records from 90 subjects. Each record contains the raw noisy ECG signal and the filtered version that represents the clean signal. The ECG signal is often corrupted by power line interference, contact noise and motion artefacts. We choose 217 ECG records (70\% of the sequences) for training,  62 (20\%) for validation  and 31 (10\%) for testing. The total number of samples is around 385k. We use the training set for generating the synthetic tasks, where in each task, we add Gaussian noise to the clean training samples, as in $\mathbf{x} = h_{\tau = \mathcal{N}(\mu,\sigma)}(\mathbf{y})$, where $h_{\tau}(.)$ is defined by $\mathbf{x}(j) = \mathbf{y}(j) + \eta(j)$ for $j\in \{1,...,D\}$, $\tau$ is represented by a Gaussian noise $\eta(j) \sim \mathcal{N}(\mu,\sigma)$, with $\mu \sim U(-0.1,0.1)$ and standard deviation $\sigma \sim U(0,0.3)$ for $U(a,b)$ denoting uniform distribution in $[a,b]$ and $\mathcal{N}(.)$ representing the Gaussian distribution with mean $\mu$ and variance $\sigma$. We generate the training samples per task, where each sample is represented by a window of $D=30$ consecutive signal measurements with stride one. To perform fine-tuning for transfer learning or meta-learning, we randomly draw $k$ samples from the test set and use the rest for evaluation.

\noindent \textbf{Model.} The network is a denoising autoencoder~\cite{casas2018adversarial}, consisting of 3 encode and 3 decoder linear layers with 150 units, and a latent representation layer with 25 units~\cite{casas2018adversarial}. The activation function is ReLU, which is applied in all layers, except the last one that uses linear activation. The input and output for the model consists of a 30-element vector. During meta-training, we use the AdaDelta~\cite{zeiler2012adadelta} optimization for the inner loop (i.e.~, per task optimization). AdaDelta showed a more stable performance than stochastic gradient descent. The learning rate is 1.5 and the number of epochs is 10. For the outer loop, we iterate for 800 epochs with step-size $\epsilon = 0.01$. The number of samples for few-shot evaluation is $k = 10$.

\noindent \textbf{Results.} The results are summarized in Table~\ref{eval-ECG}. We evaluate for 50 and 100 generated tasks using the Gaussian noise model. We experimentally found that a much larger number of tasks is necessary for this experiment, compared to image-based denoising. The same experiments have been conducted for supervised- and transfer-learning. We observe that increasing the number of tasks improves the performance almost by 1dB. Generating more tasks for 10-shot learning did not have a large impact on the overall performance of our model. Comparing to the other two approaches, we see that only transfer learning achieves considerable denoising and is closer to our results. We also compare our approach with a non-learning algorithm, as with the earlier experiments, where we evaluate a Wavelets approach~\cite{tagkey2009iii}, resulting in -5.90dB, which is much worse than our approach. Here, learning-based denoising produces more accurate results than non-learning approaches.

\begin{table}
\centering
\begin{tabular}{|l|c|c|}
\hline
            & \# Tasks & SNR(db)\\ \hline
Initial Noise       &         -       &   -6.16   \\ \hline
\multirow{2}{*}{\shortstack[l]{Supervised\\ Learning}} &     50     & -5.80 \\ \cline{2-3}
                              &     100    & -5.45 \\ \hline
\multirow{2}{*}{\shortstack[l]{Transfer\\ Learning}} &     50     & 0.53 \\ \cline{2-3}
                              &     100    & 2.82 \\ \hline
\multirow{2}{*}{\shortstack[l]{Meta-Denoising}} &     50     & 2.00 \\ \cline{2-3}
                              &     100    & \textbf{2.97} \\ \hline
\end{tabular}
\caption{We evaluate our approach on ECG sequences. The supervised-learning is trained on the same data as our approach. The transfer-learning has been fine-tuned on the same k-shot data that we use during fine-tuning. The evaluation metric is the signal-to-noise ratio (SNR), larger is better. Best result is shown in bold font. \vspace{+.1in}}
\label{eval-ECG}
\end{table}

\subsection{Statistical Significance}
\label{sec:statistical_significance}

To examine the statistical significance of our approach, we conduct one-tailed paired t-test for all experiments, comparing our approach to supervised and transfer learning. For all experiments, the test results show that under the significance level of $10^{-3}$, the performance of our approach is always significantly better. 

\subsection{K-shot Learning}
We further study the performance of our approach under different k-shots of learning. The goal is to observe the performance drop when reducing the number fine-tuning samples $k$, which is a common experiment for few-shot classification. We selected as reference the CT-Scan data set with eight Poisson tasks, since this model achieve the best performance as it is shown in Table~\ref{eval-CT2}. We have trained this approach for 1-, 3-, 5-, 7- and 10-shot learning, and followed the same protocol for the transfer learning. Table~\ref{eval-kshot-ct} summarizes the results. It is clear that larger values for $k$ improves the results. Interestingly, we are able to denoise even with 3-shot learning. The same conclusion holds for transfer learning, where we chose the model with four Poisson tasks as reference. By comparing the two algorithms, we see that our method always reaches higher PSNR for both dose tube currents.

\subsection{Meta-Learning with MAML}
We have considered MAML~\cite{finn2017model} as a second meta-learning algorithm for the few-shot denoising problem. We have observed though that it is less stable during training and significantly slower in terms of convergence. MAML training has been at least four times slower than Reptile independently of the amount of generated tasks. Finally, it does not generalize as well as Reptile. In particular, we have again considered the CT-Scan data set where MAML performs at best when using eight Gaussian tasks. The PSNR is 35.08 for $5\%$ tube current and 37.68 for for $10\%$ tube current. These results are far behind Reptile results in Table~\ref{eval-CT2}.

\begin{table}[]
\centering
\begin{tabular}{|l|c|c|c|}
\hline
                              & k-shot &10\%                    &5\%                   \\\hline
Initial Noise                 & -   &          38.28      &    35.18                  \\ \hline
\multirow{3}{*}{\shortstack[l]{Transfer Learning}} & 1 &   36.82    &   34.94        \\ \cline{2-4} 
                             & 3 &   38.43    &   36.21         \\ \cline{2-4}
                              &  5&       39.28       &   36.79          \\ \cline{2-4}
                              &  7&       38.87       &   36.88         \\ \cline{2-4}
                              &  10&       38.98       &  36.96          \\ \hline
\multirow{3}{*}{\shortstack[l]{Meta-Denoising}}  & 1 &  35,94     &  33,82               \\ \cline{2-4}
                                & 3 &    38.90   &             36.90      \\ \cline{2-4} 
                              & 5&       39.30       &             37.40          \\ \cline{2-4}
                              & 7&       39.72       &            37.87           \\ \cline{2-4}
                              & 10&       \textbf{39.99}       &             \textbf{37.96}          \\ \hline                              

\end{tabular}
\caption{K-shot learning experiment. We evaluate our approach and transfer learning for 1-, 3-, 5-, 7- and 10-shot learning on the CT-Scan database. The evaluation metric is PSNR. Best result per column is shown in bold font.}
\label{eval-kshot-ct}
\end{table}

\section{Conclusion}
We have introduced few-shot meta-learning for denoising. Our approach applies gradient-based meta-learning to signal denoising based on synthetically generated tasks. In our evaluation, we explored two image- and a sequence-based problems, where we demonstrate that: 1) meta-learning is effective given sufficient amount of synthetic tasks; 2) meta-learning works better than supervised- and transfer-learning under the assumptions listed in Sec.~\ref{sec:introduction}; 3) meta-learning can produce competitive denoising results, compared with a fully supervised training with real data; 4) it is more accurate than non-learning approaches, when the synthetic noise represents well the (latent) noise model used to form the real noisy image; and 5) learning-based denoising inference is significantly faster than non-learning models. Our study provides strong indication that meta-learning has the potential to become the standard learning algorithm for denoising.

{\small
\bibliographystyle{ieee_fullname}
\bibliography{engbib}
}

\end{document}